\documentclass{elsart} 

\newif\ifignore % when set to true, additional text appears containing
\ignorefalse

\usepackage{etex} % to get more space
\usepackage{amsmath}
\usepackage{amssymb}
\usepackage{amstext}
\usepackage{mathtools}
\usepackage{mathrsfs}
\usepackage{stmaryrd}
\usepackage{xspace}
\usepackage{enumitem}
\usepackage{cmll}
\usepackage{hyperref}
\usepackage{url}
\usepackage{listings}
\usepackage{fancybox}

\usepackage{xypic}
\usepackage[all,cmtip]{xy}
\xyoption{2cell}
\UseTwocells
\xyoption{tips}
\usepackage{tikz}
\usetikzlibrary{circuits.ee.IEC}

\newenvironment{proof}{\begin{trivlist} \item[\hskip \labelsep%
{\bf Proof.}]}{\end{trivlist}}

\renewcommand{\arraystretch}{1.3}
\newcommand{\QEDbox}{\square}

\newcommand{\klafter}{\mathbin{\ast}}
\DeclareSymbolFont{T1op}{T1}{cmr}{m}{n}
\SetSymbolFont{T1op}{bold}{T1}{cmr}{bx}{n}
\DeclareMathSymbol{\mathguilsinglleft}{\mathopen}{T1op}{'016}
\DeclareMathSymbol{\mathguilsinglright}{\mathclose}{T1op}{'017}

\newcommand{\idmap}[1][]{\ensuremath{\mathrm{id}_{#1}}}

\renewcommand{\tt}{\ensuremath{\mathit{tt}}}
\newcommand{\ff}{\ensuremath{\mathit{ff}}}

\newcommand{\ket}[1]{\ensuremath{|{\kern.1em}#1{\kern.1em}\rangle}}
\newcommand{\bigket}[1]{\ensuremath{\big|{\kern.1em}#1{\kern.1em}\big\rangle}}

\newcommand{\one}{\ensuremath{\mathbf{1}}}

\newcommand{\andthen}{\mathrel{\&}}

\newcommand{\Prob}{\mathop{\mathsf{P}}}

\newsavebox\sbpto
\savebox\sbpto{\begin{tikzpicture}[baseline=-2.5pt]
            \filldraw[fill=white,draw=white] circle (1.4pt);
            \filldraw[fill=white,draw=black,line width=0.2pt] circle
(1.2pt);
                \end{tikzpicture}}
\newcommand\pto{\mathrel{\ooalign{$\to$\cr
            \hfil\!$\usebox\sbpto$\hfil\cr}}}
            
\newcommand\kto[2]{#1 \pto #2}

\newcommand{\distributionsymbol}{\mathcal{D}}

\newcommand{\Dst}{\distributionsymbol}

\newcommand{\Giry}{\mathcal{G}}

\newcommand{\UF}{\ensuremath{\mathcal{U}{\kern-.75ex}\mathcal{F}}}

\newcommand{\Kl}{\mathcal{K}{\kern-.4ex}\ell}
\newcommand{\EM}{\mathcal{E}{\kern-.4ex}\mathcal{M}}

\newcommand{\Ef}{\ensuremath{\mathcal{E}{\kern-.5ex}f}}
\newcommand{\intd}{{\kern.2em}\mathrm{d}{\kern.03em}}

\newcommand{\OF}{\ensuremath{\mathcal{O}{\kern-.1em}\mathcal{F}}}
\newcommand{\Closed}{\ensuremath{\mathcal{C}{\kern-.05em}\ell}}

\newcommand{\smoke}{\textsf{smoke}}
\newcommand{\asia}{\textsf{asia}}
\newcommand{\tub}{\textsf{tub}}
\newcommand{\lung}{\textsf{lung}}
\newcommand{\bronc}{\textsf{bronc}}
\newcommand{\either}{\textsf{either}}
\newcommand{\dysp}{\textsf{dysp}}
\newcommand{\xray}{\textsf{xray}}

\makeatother
 \newdir{ >}{{}*!/-7.5pt/@{>}}
 \newdir{|>}{!/4.5pt/@{|}*:(1,-.2)@^{>}*:(1,+.2)@_{>}}
 \newdir{ |>}{{}*!/-3pt/@{|}*!/-7.5pt/:(1,-.2)@^{>}*!/-7.5pt/:(1,+.2)@_{>}}

\newsavebox\sbground
\savebox\sbground{\begin{tikzpicture}[circuit ee IEC,yscale=0.5,xscale=0.4]
                \draw (0,-2ex) to (0,0) node[ground,rotate=90,xshift=.65ex] {};
                \end{tikzpicture}}

\DeclareFixedFont{\ttb}{T1}{txtt}{bx}{n}{11} % for bold
\DeclareFixedFont{\ttm}{T1}{txtt}{m}{n}{11}  % for normal
\usepackage{color}
\definecolor{deepblue}{rgb}{0,0,0.5}
\definecolor{deepred}{rgb}{0.6,0,0}
\definecolor{deepgreen}{rgb}{0,0.5,0}
\definecolor{lightgray}{rgb}{0.83,0.83,0.83}

\newcommand\pythonstyle{\lstset{
backgroundcolor = \color{lightgray},
language=Python,
basicstyle=\ttm,
otherkeywords={self,>>>,...},             % Add keywords here
keywordstyle=\ttb\color{deepblue},
emph={MyClass,__init__},          % Custom highlighting
emphstyle=\ttb\color{deepred},    % Custom highlighting style
stringstyle=\color{deepgreen},
frame=tb,                         % Any extra options here
showstringspaces=false            % 
}}

\lstnewenvironment{python}[1][]
{
\pythonstyle
\lstset{#1}
}
{}

\newcommand\pythoninline[1]{{\pythonstyle\lstinline!#1!}}

\begin{document}
\begin{frontmatter}
\title{A Channel-based Exact Inference Algorithm for Bayesian 
   Networks\thanksref{ERC}}

\thanks[ERC]{The research leading to these results has received
  funding from the European Research Council under the European
  Union's Seventh Framework Programme (FP7/2007-2013) / ERC grant
  agreement nr.~320571}

\author{Bart Jacobs}

\address{Institute for Computing and Information Sciences, \\
Radboud University Nijmegen\\ 
P.O. Box 9010, 6500 GL Nijmegen, The Netherlands \\
Email: {\tt B.Jacobs@cs.ru.nl}\quad
URL: {\tt http://www.cs.ru.nl/B.Jacobs}
}

%\date{}

\begin{abstract}
This paper describes a new algorithm for exact Bayesian inference that
is based on a recently proposed compositional semantics of Bayesian
networks in terms of channels. The paper concentrates on the ideas
behind this algorithm, involving a linearisation (`stretching') of the
Bayesian network, followed by a combination of forward state
transformation and backward predicate transformation, while evidence
is accumulated along the way.  The performance of a prototype
implementation of the algorithm in Python is briefly compared to a
standard implementation (\texttt{pgmpy}): first results show
competitive performance.
\end{abstract}

\end{frontmatter}

%\tableofcontents

\section{Introduction}\label{IntroSec}

% Sources:
%
% https://people.eecs.berkeley.edu/~wainwrig/Talks/Wainwright_PartI.pdf
% 
% https://www.quora.com/Probabilistic-graphical-models-what-are-the-relationships-between-sum-product-algorithm-belief-propagation-and-junction-tree-algorithm
%
% https://www.cs.cmu.edu/~epxing/Class/10708-14/scribe_notes/scribe_note_lecture4.pdf

In general, inference is about answering probabilistic queries of the
form: given this-and-this as evidence, what is the likelihood of that?
The focus in this paper is on \emph{exact} inference, where precise
answers are sought and not approximations. In general, inference is
computationally very expensive. In probabilistic graphical models one
can exploit the graph structure in various ways. Here we concentrate
on Bayesian networks, whose underlying graphs are directed and
acyclic.

This paper presents a new algorithm for (exact) inference in Bayesian
networks. The algorithm is based on a novel logical perspective on
Bayesian networks, which has its roots in category theory and in the
semantics of programming languages. Category theory is a mathematical
formalism that concentrates on universal properties of mathematical
constructs, see \textit{e.g.}~\cite{Awodey06,Pierce91,MacLane71}. In
computer science it is especially used to capture the essential
structural properties in type theory and programming semantics. Its
approach gives a `helicopter' view in which, for instance, the
similarities between discrete and continuous probabilistic computation
(or even quantum computation) are emphasised~\cite{Jacobs17a}. In all
these cases one deals with symmetric monoidal categories, for which a
graphical language (`string diagrams', see
\textit{e.g.}~\cite{Selinger11}), has been developed.  This language
exploits the compositional character of this setting, with sequentical
composition $\klafter$ interacting appropriately with parallel
composition $\otimes$. This string diagram calculus is similar, but
subtly different from the graphical languages used for Bayesian
networks, see \textit{e.g.}~\cite{Fong12,JacobsZ16,JacobsZ18}.

Here we shall not go into these underlying categorical theories nor
into string diagrams. We refer to~\cite{JacobsZ18} for a recent
(gentle) introduction. Instead, we present our approach in a rather
hands-on manner, via a well-known Bayesian network example from the
literature. It is based on the notion of channel, as abstraction of
stochastic matrix. In fact, it is shown how conditional probability
tables in Bayesian networks can be naturally seen as channels. This is
not a novelty, but what is crucial for the current approach, is that
along a channel one can do forward transformation $\gg$ of states
(distributions), and backward transformation $\ll$ of predicates, see
also~\cite{Jacobs17b}. These ideas are explained in
Sections~\ref{sec:baynet} and~\ref{sec:predicates}. There it is also
shown how updating of states with predicates (evidence), in
combination with forward and backward transformations, allows us to do
Bayesian inference at a high level of abstraction, essentially by
following the graph structure. We are well aware of the fact that this
channel-based formalism deviates from the standard notation and
terminology in Bayesian probability.  But we do hope that the new
algorithm for Bayesian inference that builds on this approach provides
enough motivation to look into it.

Indeed, the interpretation of a Bayesian network in terms of channels
is essential for the new inference algorithm, which is explained in
Section~\ref{sec:algo}. The algorithm consists of three steps: (1)~the
Bayesian network is stretched to a linear chain of channels;
(2a)~state transformation is performed forwardly, from the beginning
of this chain, to the observation point in the chain, while evidence
is incorporated along the way; (2b)~predicate transformation is
performed backwardly, from the end of the chain, to the observation
point, while evidence is accumulated along the way. Finally, at the
observation point, the (relevant marginal of the) updated state is
returned as output. Section~\ref{sec:algo} provides more details,
including two optimisation steps.

A prototype implementation of this stretch-and-infer algorithm has
been written in Python, simply in order to investigate whether the
approach works and can handle non-trivial examples. The implementation
builds on the EfProb library~\cite{ChoJ17b} for channel-based
probabilistic computations. This paper concentrates on the methodology
to use channel-based compositional semantics for Bayesian inference
--- the main intellectual contribution --- and not on this prototype
implementation. Nevertheless, a brief comparison with the standard
Variable Elimination algorithm (see
\textit{e.g.}~\cite{KollerF09,Barber12,JensenN07}) is given in
Subsection~\ref{subsec:comparison}. The comparison is complicated by
the level of non-determinism involved in these inference
algorithms. Nevertheless, the comparison does indicate that our
algorithm can handle non-trivial (standard) examples faster than
Variable Elimination in the widely used Python library \texttt{pgmpy}
for probabilistic graphical modeling~\cite{AnkanP15}.

The computations that have to be performed for inference are in
principle well-known, by Bayes' rule, namely suitable sums and
products, and also divisions for normalisation. The cleverness in
various algorithms lies in the order in which these computations are
performed. In the terminology of Variable Elimination, this means the
order in which variables are eliminated.  The analogue in our
algorithm is the order in which channels are lined-up in a chain,
during the first, stretching phase of the computation. This order is
randomly produced in a non-deterministic process, through iterations
over \emph{sets}, instead of \emph{lists}, of ancestors of
nodes. However, what we can do is perfom `dry runs' of the stretching
in order to find the best order (with the least width), as
optimisation of our algorithm, see Subsection~\ref{subsubsec:dryrun}.

The outcome of an algorithm should not depend on its non-deterministic
choices. In our algorithm this leads to the mathematical question
whether the outcome is independent of the ordering of channels in a
chain.  Section~\ref{sec:correct} is devoted to answering this
question, positively, by reducing the situation to its essential
form. Finally, Section~\ref{sec:conclusions} describes future work.

\section{The channel-based approach to Bayesian networks}\label{sec:baynet}

\begin{figure}
$$\xymatrix@C-1pc@R-0.5pc{
\ovalbox{\strut smoke}\ar[dr]\ar[ddd]
\llap{\smash{\setlength\tabcolsep{0.2em}\renewcommand{\arraystretch}{1}
\begin{tabular}{|c|}
\hline
$\Prob(\text{smoke})$ \\
\hline\hline
$0.5$ \\
\hline
\end{tabular}}\hspace*{6em}}
& &
\ovalbox{\strut asia}\ar[d]
\rlap{\hspace*{2em}\smash{\setlength\tabcolsep{0.2em}\renewcommand{\arraystretch}{1}
\begin{tabular}{|c|}
\hline
$\Prob(\text{asia})$ \\
\hline\hline
$0.01$ \\
\hline
\end{tabular}}}
\\
& \hspace*{-0.5em}\ovalbox{\strut lung}\ar[ddr]
\llap{\smash{\setlength\tabcolsep{0.2em}\renewcommand{\arraystretch}{1}
\begin{tabular}{|c|c|}
\hline
smoke & $\Prob(\text{lung})$ \\
\hline\hline
$t$ & $0.1$ \\
\hline
$f$ & $0.01$ \\
\hline
\end{tabular}}\hspace*{7em}}
& 
\ovalbox{\strut tub}\ar[dd]
\rlap{\hspace*{1.5em}\smash{\setlength\tabcolsep{0.2em}\renewcommand{\arraystretch}{1}
\begin{tabular}{|c|c|}
\hline
asia & $\Prob(\text{tub})$ \\
\hline\hline
$t$ & $0.05$ \\
\hline
$f$ & $0.01$ \\
\hline
\end{tabular}}}
\\
\\
\ovalbox{\strut bronc}\ar[dd]
\llap{\smash{\setlength\tabcolsep{0.2em}\renewcommand{\arraystretch}{1}
\begin{tabular}{|c|c|}
\hline
smoke & $\Prob(\text{bronc})$ \\
\hline\hline
$t$ & $0.6$ \\
\hline
$f$ & $0.3$ \\
\hline
\end{tabular}}\hspace*{5em}}
& &
\ovalbox{\strut either}\ar[dd]\ar[lldd]
\rlap{\hspace*{1.5em}\smash{\setlength\tabcolsep{0.2em}\renewcommand{\arraystretch}{1}
\begin{tabular}{|c|c|c|}
\hline
lung & tub & $\Prob(\text{either})$ \\
\hline\hline
$t$ & $t$ & $1$ \\
\hline
$t$ & $f$ & $1$ \\
\hline
$f$ & $t$ & $1$ \\
\hline
$f$ & $f$ & $0$ \\
\hline
\end{tabular}}}
\\
\\
\ovalbox{\strut dysp}
\llap{\smash{\setlength\tabcolsep{0.2em}\renewcommand{\arraystretch}{1}
\begin{tabular}{|c|c|c|}
\hline
bronc & either & $\Prob(\text{dysp})$ \\
\hline\hline
$t$ & $t$ & $0.9$ \\
\hline
$t$ & $f$ & $0.7$ \\
\hline
$f$ & $t$ & $0.8$ \\
\hline
$f$ & $f$ & $0.1$ \\
\hline
\end{tabular}}\hspace*{5em}}
& &
\ovalbox{\strut xray}
\rlap{\hspace*{1.5em}\smash{\setlength\tabcolsep{0.2em}\renewcommand{\arraystretch}{1}
\begin{tabular}{|c|c|}
\hline
either & $\Prob(\text{xray})$ \\
\hline\hline
$t$ & $0.98$ \\
\hline
$f$ & $0.05$ \\
\hline
\end{tabular}}}
\\
\\
}$$
\caption{The `Asia' example from \url{bnlearn.com} with its
  conditional probability tables}
\label{fig:asiacpt}
\end{figure}
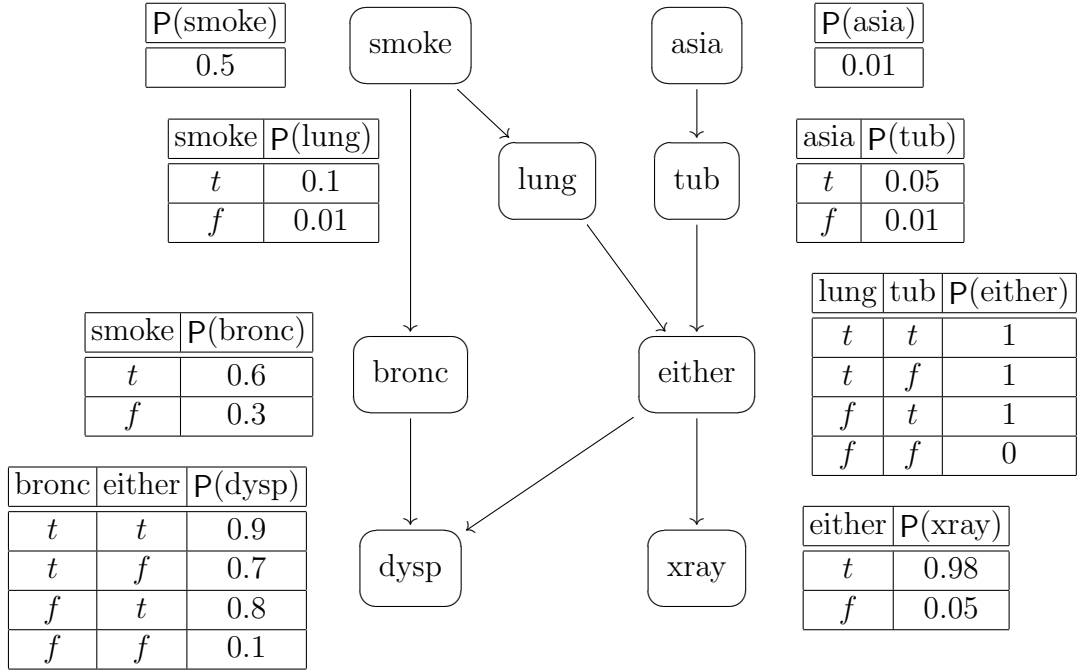

This section explains in a concrete way how Bayesian networks can be
described conveniently in terms of states $\omega$, channels $c$, and
forward transformation $c \gg \omega$ of a state along a channel. For
a more elaborate introduction we refer to~\cite{JacobsZ18}, and
to~\cite{Jacobs15d} for a more abstract account of the underlying
category theory. We shall use the (categorical) calculus of channels,
with sequential $\klafter$ and parallel $\otimes$ composition, in
order to calculate marginals of a Bayesian network. We use as running
example the standard `Asia' Bayesian network, as reproduced in
Figure~\ref{fig:asiacpt}. It describes a simple network for medical
diagnostics.

\subsection{Sequential composition}\label{subsec:sequential}

We write $t,f$ for the two `true' and `false' elements of the
2-element set $2 = \{t,f\}$. The initial nodes smoke and asia in
Figure~\ref{fig:asiacpt} come equipped with two probability
distributions on this set $2$. We like to write them as:
\[ \begin{array}{rclcrcl}
\smoke
& = &
0.5\ket{t} + 0.5\ket{f}
& \qquad\mbox{and}\qquad &
\asia
& = &
0.01\ket{t} + 0.99\ket{f}.
\end{array} \]
\noindent More generally, a probability distribution on an $n$-element
set $X = \{x_{1}, \ldots, x_{n}\}$ will be written as formal convex
sum $\sum_{i} r_{i}\ket{x_i}$ with $r_{i}\in [0,1]$ satisfying
$\sum_{i}r_{i}=1$.  Such a distribution will often be called a
\emph{state}. We write $\Dst(X)$ for the set of such
distributions/states on $X$. Thus we have $\smoke, \asia \in\Dst(2)$.

A conditional probability table (CPT) for a node in
Figure~\ref{fig:asiacpt} with $n$ parents corresponds to a function
$2^{n} \rightarrow \Dst(2)$. For instance, we can read off:
\[ \begin{array}{ccc}
\begin{array}{rcl}
\lung(t) & = & 0.1\ket{t} + 0.9\ket{f}
\\
\lung(f) & = & 0.01\ket{t} + 0.99\ket{f}
\end{array}
& \hspace*{4em} &
\begin{array}{rcl}
\dysp(t,t) & = & 0.9\ket{t} + 0.1\ket{f}
\\
\dysp(t,f) & = & 0.7\ket{t} + 0.3\ket{f}
\\
\dysp(f,t) & = & 0.8\ket{t} + 0.2\ket{f}
\\
\dysp(f,f) & = & 0.1\ket{t} + 0.9\ket{f}
\end{array}
\end{array} \]
\noindent More generally, for sets $X,Y$, a \emph{channel} from $X$ to
$Y$ is a function $X \rightarrow \Dst(Y)$. It consists of an
$X$-indexed collection of states on $Y$. It may be understood as a
$\#X\times\#Y$ stochastic matrix. We often write $c\colon X\pto Y$
when we mean that $c$ is a channel from $X$ to $Y$, and thus a
function from $X$ to $\Dst(Y)$. As we shall see, this special arrow
$\pto$ is convenient when we compose channels, both sequentially and
in parallel. Notice by the way that a state/distribution on a set $X$
is in fact also a channel, namely a channel of the form $1\pto X$,
where $1 = \{0\}$ is a singleton set.

A very basic operation associated with channels is \emph{state
  transformation}. If we have a state $\omega =
\sum_{i}r_{i}\ket{x_i}$ on $X$, and a channel $c\colon \kto{X}{Y}$, we
can form a new state on $Y$, written as $c \gg \omega$. If we write
$c(x_{i}) = \sum_{j} s_{ij}\ket{y_{ij}}$, then:
\[ \begin{array}{rcl}
c \gg \omega
& \coloneqq &
\displaystyle\sum_{ij} r_{i}\cdot s_{ij} \ket{y_{ij}}.
\end{array} \]
\noindent For instance, in the context of the Asia example,
\[ \begin{array}{rcl}
\lung \gg \smoke
& = &
0.055\ket{t} + 0.9450\ket{f}
\\
\tub \gg \asia
& = &
0.0104\ket{t} + 0.9896\ket{f}.
\end{array} \]
\noindent Thus, with state transformation $\gg$ we can calculate
`marginal' states for nodes in the graph, by just following the path.
So far we have done single steps, we can also do this for multiple
steps by composing channels sequentially. For channels $c\colon
\kto{X}{Y}$ and $d\colon \kto{Y}{Z}$ we can form a composite channel
$d\klafter c \colon \kto{X}{Z}$ in the following way. Recall that we
have to define $d\klafter c$ as a function $X \rightarrow \Dst(Z)$. 
We use state transformation along the channel $d$ in:
\[ \begin{array}{rcl}
\big(d \klafter c\big)(x)
& \coloneqq &
d \gg c(x).
\end{array} \]
\noindent It is not hard to see that this composition is associative.
There is an identity channel $\idmap\colon \kto{X}{X}$ given by
$\idmap(x) = 1\ket{x}$, so that $\idmap \klafter c = c = c \klafter
\idmap$. Further, one has `functoriality' properties:
\[ \begin{array}{rclcrcl}
(d \klafter c) \gg \omega
& = &
d \gg (c \gg \omega)
& \qquad\qquad &
\idmap \gg \omega
& = &
\omega.
\end{array} \]

We summarise the situation at some higher level of
abstraction. Roughly, a Bayesian network consists of a directed
acyclic graph (a `DAG') with a conditional probability table for each
node. These tables have to be of the right `dimension'. Associated
with each node $A$ of the network, a finite set $\underline{A}$ is
implicitly associated, with at least two elements. We write $\# A$ for
the number of elements of $\underline{A}$. Often, $\# A = 2$, as in
the Asia example, but larger sets are allowed. If the node $A$ has
(direct) parent nodes $B_{1}, \ldots, B_{n}$, then the conditional
probability table associated with $A$ has $\#B_{1} \cdot \ldots \cdot
\#B_{n}$ many distributions on $\underline{A}$. Thus we can associate
such a node $A$ with a channel $\underline{B}_{1}\times \cdots \times
\underline{B}_{n} \pto \underline{A}$. Channels can be composed
sequentially, and states/distributions can be transformed along
channels.

\subsection{Parallel composition}\label{subsec:parallel}

In a Bayesian network like in Figure~\ref{fig:asiacpt} certain
structure occurs in parallel. For instance, the two initial nodes
smoke and asia can be combined into a joint distribution
$\smoke\otimes\asia$ on $2\times 2$. We now describe how such parallel
composition $\otimes$ works, both for states and for channels.

Given two states $\omega = \sum_{i}r_{i}\ket{x_i}$ on $X$ and $\rho = 
\sum_{j}s_{j}\ket{y_i}$ on $Y$, we can form the `joint' or `product'
state $\omega\otimes\rho$ on $X\times Y$, namely:
\[ \begin{array}{rcl}
\omega \otimes \rho
& \coloneqq &
\displaystyle\sum_{i,j} r_{i}\cdot s_{j} \ket{x_{i},y_{j}}.
\end{array} \]
\noindent For instance,
\[ \begin{array}{rcl}
\lefteqn{\smoke \otimes \asia}
\\
& = &
0.005\ket{t,t} + 0.495\ket{t,f} + 0.005\ket{f,t} + 0.495\ket{f,f}
\\
\lefteqn{(\lung \gg \smoke) \otimes \asia} 
\\
& = &
0.00055\ket{t,t} + 0.0544\ket{t,f} + 0.00945\ket{f,t} + 0.936\ket{f,f}.
\end{array} \]
\noindent We extend $\otimes$ from states to channels via a pointwise
definition: for $c\colon \kto{X}{Y}$ and $d \colon \kto{A}{B}$ we
get $c\otimes d \colon \kto{X\times A}{Y\times B}$ via:
\[ \begin{array}{rcl}
\big(c\otimes d\big)(x,a)
& \coloneqq &
c(x) \otimes d(a).
\end{array} \]
\noindent With parallel composition $\otimes$ we can compute more
marginals in Figure~\ref{fig:asiacpt}. For instance, the marginal at
the `either' node can be obtained in several equivalent ways as:
\[ \begin{array}{rcl}
\lefteqn{\big( \either \klafter (\lung \otimes \tub)\big) \gg 
   (\smoke \otimes \asia)}
\\
& = &
\either \gg ((\lung \otimes \tub) \gg (\smoke \otimes \asia))
\\
& = &
\either \gg \big((\lung \gg \smoke) \otimes (\tub \gg \asia)\big)
\\
& = &
0.0648\ket{t} + 0.935\ket{f}.
\end{array} \]
\noindent In order to cover the whole network in this way we need to
have copy (and projection) channels. We write $\Delta\colon
\kto{X}{X\times X}$ for the channel given by $\Delta(x) =
1\ket{x,x}$. Similarly, there are projection channels $\pi_{i} \colon
\kto{X_{1}\times X_{2}}{X_i}$, for $i=1,2$, given by
$\pi_{i}(x_{1},x_{2}) = 1\ket{x_{i}}$. State transformation $\pi_{i}
\gg \omega$ corresponds to \emph{marginalisation}.

\begin{figure}
\begin{center}
\raisebox{1em}{\includegraphics[scale=0.5]{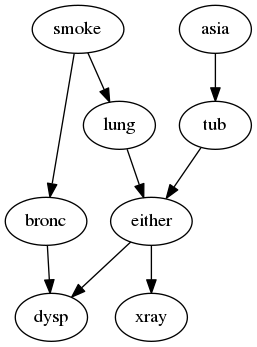}}
\hspace*{3em}
\includegraphics[scale=0.4]{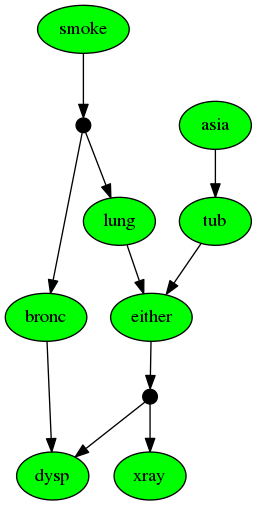}
\end{center}
\caption{The underlying graph of the Asia example from
  Figure~\ref{fig:asiacpt}, in standard form on the left, and with
  explicit copy nodes $\bullet$ on the right.}
\label{fig:asiabn}
\end{figure}

There is a convention in Bayesian networks that if a node has multiple
outgoing arrows, then the same data is sent over all these wires. This
means that there is implicit copying. We need to make this explicit,
for a compositional description of Bayesian networks, and use slightly
non-standard notation with copying explicit, as on the right in
Figure~\ref{fig:asiabn}.

With this explicit copying in place we see how calculate the marginal
for the `dysp' node. It arrises as composition:
\[ \begin{array}{rcl}
\lefteqn{\big(\dysp \klafter (\idmap \otimes \either) \klafter 
   (\bronc \otimes \lung \otimes \idmap) \klafter 
   (\Delta \otimes \tub)\big) \gg (\smoke \otimes \asia)}
\\
& = &
0.3975\ket{t} + 0.6025\ket{f}.
\end{array}\hspace*{20em} \]
This state transformation formulation gives a systematic approach to
computing marginals, basically by following the graph structure, and
translating it into a corresponding channel expression. However,
computing the outcome by hand is painful. The EfProb
tool~\cite{ChoJ17b} has been designed to evaluate such expressions.
Without going into details of the EfProb language, we hope the reader
sees the correspondence between the above mathematical formula and the
EfProb (Python) expression below.
\begin{python}
>>> dysp * (id @ either) * (bronc @ lung @ id) * (copy @ tub) 
...     >> (smoke @ asia)
0.3975|t> + 0.6025|f>
\end{python}
We challenge the interested reader to write down the channel
expression for computing the state for the node xray, and
also for the joint state of dysp and xray. One can use the
laws:
\[ \begin{array}{rcl}
(d \otimes e) \klafter (c \otimes f)
& = &
(d \klafter c) \otimes (e \klafter f)
\\
(c\otimes d) \gg (\omega\otimes\rho)
& = &
(c \gg \omega) \otimes (d \gg \rho).
\end{array} \]

\section{Evidence in Bayesian networks}\label{sec:predicates}

A typical Bayesian inference question, for the Asia example is: given
that a patient has been in Asia and has a positive xray, what is the
likelihood of having dyspnea? In this question we have certain
`evidence', namely `having been in asia' and `positive xray', and we
have a specific `observation' node (namely dysp) whose resulting state
we like to infer. This section introduces the machinery to answer such
questions under the channel-based interpretation of Bayesian networks.

We take a quite general perspective on evidence, namely using the
`soft' or `uncertain' view (using terminology from~\cite{Barber12}).
Also we use `predicate' instead of `evidence' in line with the
terminology from (logical) program semantics, see~\cite{JacobsZ18}.

Quite generally, a predicate $p$ on a set $X$ is a function $p\colon X
\rightarrow [0,1]$. Thus, the truth value $p(x) \in [0,1]$ associated
with an element $x\in X$ involves uncertainty. If either $p(x) = 0$ or
$p(x) = 1$, for each $x\in X$, the predicate $p$ is called
\emph{sharp}. A sharp predicate is also called an event. Each set $X$
carries the `truth' predicate $\one\colon X\rightarrow [0,1]$, given
by $\one(x) = 1$ for each $x\in X$. For two predicates $p,q\colon X
\rightarrow [0,1]$ we write $p\andthen q$ for the predicate obtained
by pointwise multiplication: $(p \andthen q)(x) = p(x) \cdot
q(x)$. Notice that $p\andthen \one = p = \one\andthen p$. For the
special set $2 = \{t,f\}$ we write $\tt, \ff \colon 2 \rightarrow
[0,1]$ for the (sharp) predicates with $\tt(t) = 1 = \ff(f)$ and
$\tt(f) = 0 = \ff(t)$.

\begin{defn}
\label{def:valcond}
Let $X = \{x_{1}, \ldots, x_{n}\}$ be a finite set, with a state
$\omega = \sum_{i}r_{i}\ket{x_i}$ and a predicate $p\colon X
\rightarrow [0,1]$.
\begin{enumerate}
\item We write $\omega\models p$ for the \emph{validity} of predicate
  $p$ in state $\omega$. It is defined as $\omega \models p \coloneqq
  \sum_{i} r_{i}\cdot p(x_{i})$. It is the expected value of $p$ in
  $\omega$.

\item If the validity $\omega\models p$ is non-zero we define an
  \emph{updated} state $\omega|_{p}$ on $X$, pronounced as: $\omega$
  given $p$. It is defined as:
\[ \begin{array}{rcl}
\omega|_{p}
& \coloneqq &
\displaystyle\sum_{i} \frac{r_{i}\cdot p(x_{i})}{\omega\models p} \bigket{x_i}.
\end{array} \]
\end{enumerate}
\end{defn}

As we see, the updated state $\omega|_{p}$ takes the evidence $p$ into
account via multiplication with the probabilities of $\omega$. The
resulting values have to be normalised, via the validity
$\omega\models p$, in order to obtain a distribution again. We refer
to~\cite{JacobsZ18} for more information about this updating,
including an associated version of Bayes' rule.

We need one more notion before we can start doing inference, namely
predicate transformation along a channel. Given a channel $c\colon 
\kto{X}{Y}$ and a predicate $q$ on $Y$, we can form a new predicate
$c \ll q$ on $X$, via the definition:
\[ \begin{array}{rclcrcl}
\big(c \ll q\big)(x)
& \coloneqq &
\displaystyle\sum_{j} s_{j}\cdot p(y_{j})
& \qquad\mbox{when}\qquad &
c(x)
& = &
\displaystyle\sum_{j} s_{j}\ket{y_{j}}.
\end{array} \]
Below we describe some of the mathematical laws that predicate
transformation satisfies, but first we like to illustrate how it is
used in inference. Let's ask ourselves a first question. Suppose that
we have evidence that a patient has lung cancer; what is then the
likelihood of bronchitis?

\begin{figure}
$$\xymatrix@C+1pc@R-0.5pc{ 
2\otimes 2\ar[r]^-{\Delta\otimes\idmap} &
  2\otimes 2\otimes 2\ar[d]_{\bronc\otimes\idmap\otimes\idmap} 
\\ 
& 2\otimes 2\otimes 2\ar[d]_{\idmap\otimes\lung\otimes\idmap} 
\\ 
& 2\otimes 2\otimes 2\ar[d]_{\idmap\otimes\idmap\otimes\tub} 
\\ 
& 2\otimes 2\otimes 2\ar[d]_{\idmap\otimes\either} 
\\ 
& 2\otimes 2\ar[r]^-{\idmap\otimes\Delta} 
   & 2\otimes 2\otimes 2\ar[d]^{\dysp\otimes\idmap} 
\\ 
& & 2\otimes 2\ar[d]^{\idmap\otimes\xray} 
\\ 
& & 2\otimes 2 
}$$
\caption{A stretching of the Asia Bayesian network, producing a chain
  of channels. The order of the channels with $\bronc$, $\lung$,
  $\tub$ does not matter because of Equation~\eqref{eqn:tensorshift};
  the same holds for $\dysp$ and $\xray$.}
\label{fig:asiachain}
\end{figure}
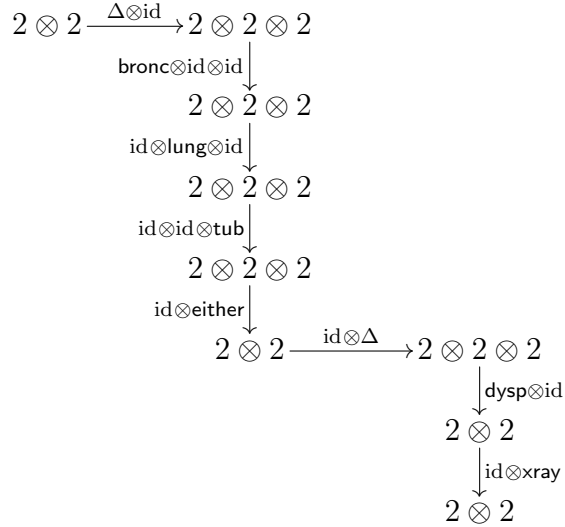

We proceed in the following methodological way, using that there is a
path from nodes `lung' to `bronc' via the initial state `smoke'. We
have the lungcancer evidence in the form of the predicate $\tt$ on the
codomain of the $\lung$ channel. We first turn it into a predicate
$\lung \ll \tt$ on its domain, and then use it to update the smoke
state to $\smoke|_{\lung \ll \tt}$. The answer is obtained via state
transformation along the $\bronc$ channel:
\[ \begin{array}{rcl}
\bronc \gg \big(\smoke|_{\lung \ll \tt}\big)
& = &
0.5727\ket{t} + 0.4273\ket{f}.
\end{array} \]
In the beginning of this section we had a more challenging question,
namely: given a visit to Asia and a negative xray, what is the
likelihood of dyspnea? For the answer we update the asia state to
$\asia|_{\tt}$ and also update the either state with predicate $\xray
\ll \ff$, and then use these updated states in the forward
recomputation of the dyspnea likelihood. In EfProb this looks as
follows --- where \pythoninline{@} is written for the tensor $\otimes$
and \pythoninline{/} for conditioning.
\begin{python}
>>> dysp >> ((bronc @ either) * (id @ lung @ tub) >> 
...  ((copy >> smoke) @ (asia / tt))) / (one @ (xray << ff))
0.3669|t> + 0.6331|f>
\end{python}
\noindent Clearly, this answer is not easy to read (and
construct). The main contributions of this paper is an algorithm for
doing this systematically, by first `stretching' a Bayesian network to
a chain of channels, as in Figure~\ref{fig:asiachain}, and then doing
forward \& backward inference along this chain. This will be described
in the next section.

But before stopping here we would like to write down some basic laws
for predicate transformation:
\[ \begin{array}{rclcrclcrcl}
(d \klafter c) \ll q
& = &
c \ll (d \ll q)
& \qquad &
\idmap \ll q
& = &
q
& \qquad &
c \ll \one
& = &
\one.
\end{array} \]
\noindent Further, the validity $c \gg \omega \models q$ is the same
as $\omega \models c \ll q$. Finally, we mention that causal reasoning
(or prediction) is given by first updating and then doing forward
state tranformation, as in $c \gg (\omega|_{p})$. In contrast,
evidential reasoning (or explanation) involves first doing predicate
transformation and then updating, as in $\omega|_{c\ll q}$. We refer
to~\cite{JacobsZ16,JacobsZ18} for further details --- where these two
approaches are called forward and backward inference.

\section{A channel-based inference algorithm}\label{sec:algo}

Having seen the basics of state transformation, predicate
transformation, and state updates, we can now proceed to the novelty
of this paper, namely a new channel-based inference algorithm for
exact Bayesian inference. The algorithm contains two steps, which we
call `stretching' and `transformation' respectively. A prototype
version of this algorithm has been developed in Python, on the basis
of the EfProb library~\cite{ChoJ17b}. It will be described as we
proceed. In the end there is a non-rigorous comparison to the standard
\emph{variable elimination} algorithm for inference, as implemented in
the widely used \texttt{pgmpy} library\footnote{See \url{pgmpy.org}
  or~\cite{AnkanP15} for more information. Belief propagation does not
  work on most of our examples because \texttt{pgmpy} fails to turn
  them into junction tree form.} for probabilistic graphical
modeling. This section will thus have the following subsections.
\begin{enumerate}
\item Stretching
\item Transformations
\item Optimisation
\item Comparison.
\end{enumerate}

\subsection{Stretching a Bayesian network}\label{subsec:stretch}

The aim of the stretching algorithm is to turn a Bayesian network into
a linear chain of channels, with one node (conditional probability
table, seen as channel) per step in the chain. Informally, the
algorithm works as follows. It first adds initial nodes to the
chain. It then adds one by one nodes (as channels in EfProb) all of
whose ancestors are already in the chain. For instance, in the Asia
example from Figure~\ref{fig:asiabn} we can add nodes/channels in a
chain in the following orders.
\begin{itemize}
\item smoke, asia, bronc, lung, tub, either, dysp, xray

\item asia, smoke, tub, lung, either, xray, bronc, dysp
\end{itemize}
\noindent Clearly, there are different possible orders in such a
`stretched' chain. When constructing a chain of channels we have to
make sure that the inputs are copied and re-arranged when
needed. Figure~\ref{fig:asiachain} describes the resulting chain of
channels corresponding to the first order given above, in which the
initial states have been omitted. Recall that we write $2 = \{t,f\}$
for the two element set.

At this stage we do not care which order is chosen
--- although we will have to say a bit more about this later in
Subsection~\ref{subsec:optim} from an optimisation perspective.  From
a mathematical perspective the order does not matter so much because
we have the following property for channel composition:
\begin{equation}
\label{eqn:tensorshift}
\begin{array}{rcl}
(c\otimes\idmap) \klafter (\idmap\otimes d)
& = &
(\idmap\otimes d) \klafter (c\otimes\idmap).
\end{array}
\end{equation}
\noindent This means that channels that do not interact can be
shifted, see Figure~\ref{fig:asiachain}. There is a bit more to say
about this when it comes to updating, but that will be postponed to
Section~\ref{sec:correct}.

This stretching algorithm has been implemented in Python. It turns a
Bayesian model formalised in the \texttt{pgmpy} library into a chain
of composable EfProb channels. The program typically yields different
outcomes for different runs. This non-determinism arises because the
program iterates over \emph{sets} (instead of \emph{lists}) of parent
nodes in a \texttt{pgmpy} model; in iterations over sets, a random
order is chosen. The implementation involves quite a bit of
bookkeeping, in order to copy nodes along the way and to swap inputs
so that they are lined up in the right order for the subsequent
channel. We are not going to describe these details here. We just like
to mention that we do the copying `lazily' in order to keep the size
of intermediate sets limited. Recall that taking a parallel product
$\otimes$ of distributions involves a multiplication of the sizes of
the underlying sets.

\begin{figure}
\begin{center}
\includegraphics[scale=0.15]{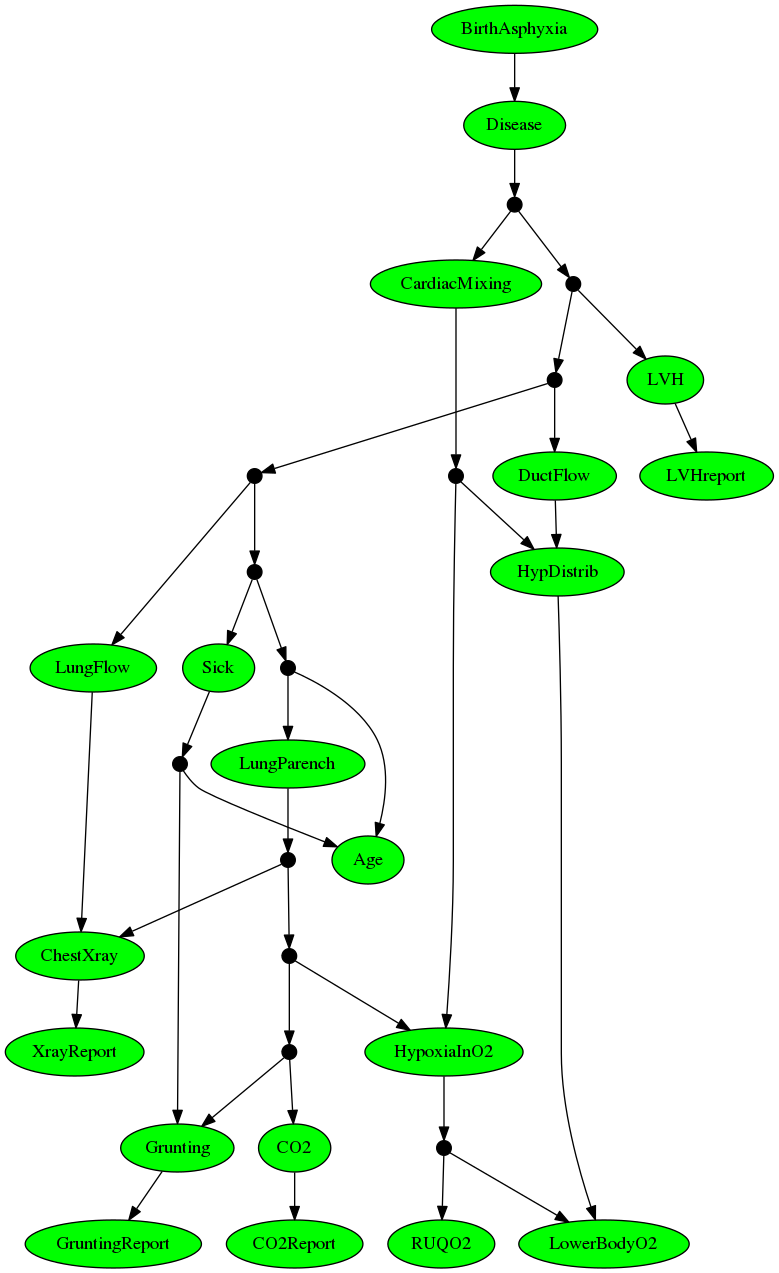}
\hspace*{0em}
\raisebox{5em}{\includegraphics[scale=0.15]{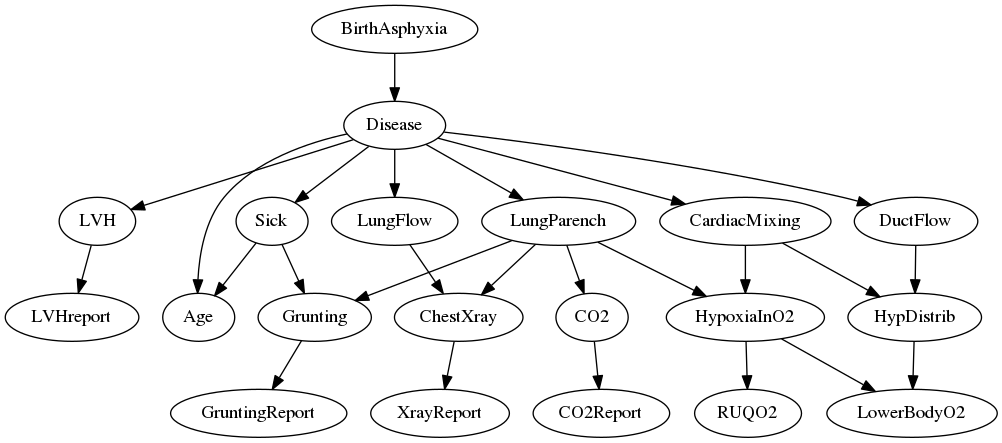}}
\hspace*{0em}
\includegraphics[scale=0.15]{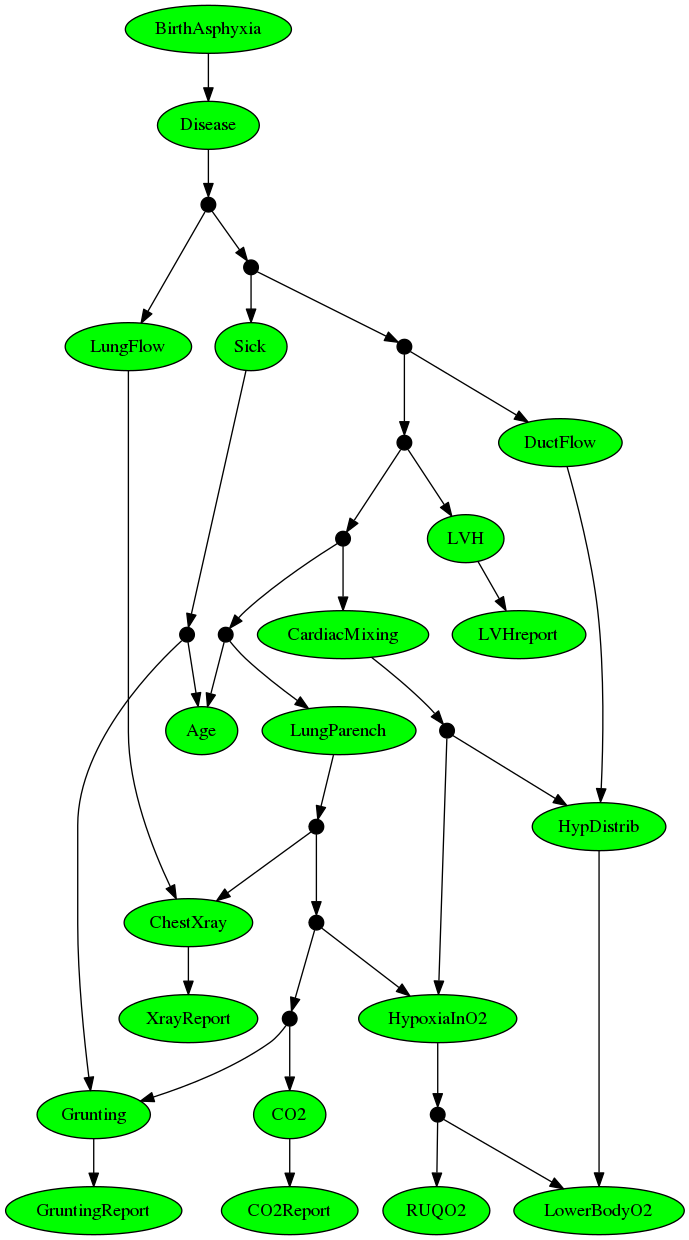}
\end{center}
\label{fig:childstretch}
\caption{The original `Child' Bayesian network from
  \texttt{bnlearn.com} in the middle, and two different ways of
  stretching it, on the left and on the right.}
\end{figure}

Figure~\ref{fig:childstretch} gives an impression of two different
outcomes of stretching the standard `Child' Bayesian network. It
clearly shows that copying is done only when needed. The graphs on the
left and right are not properly linear, but this is due to the way
such graphs are rendered (using \texttt{pydot}). The underlying
datastructures are linear chains of channels.

\subsection{Transformation along the chain of channels and updating}\label{subsec:transform}

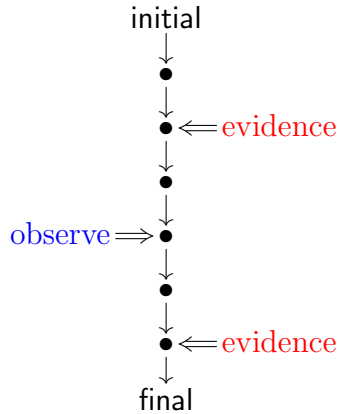
\begin{figure}
$$\xymatrix@R-1.2pc@C-1.7pc{
& \textsf{initial}\ar[d]
\\
& \bullet\ar[d]
\\
& \bullet\ar[d] & \mbox{\textcolor{red}{evidence}}\ar@{=>}[l]
\\
& \bullet\ar[d]
\\
\mbox{\textcolor{blue}{observe}}\ar@{=>}[r] & \bullet\ar[d]
\\
& \bullet\ar[d]
\\
& \bullet\ar[d] & \mbox{\textcolor{red}{evidence}}\ar@{=>}[l]
\\
& \textsf{final}
}$$
\caption{A stretched Bayesian network with the observation and
  evidence located somewhere along the chain of channels.}
\label{fig:transformation}
\end{figure}

At this stage we are ready to handle inference, as a second step in
the algorithm, after stretching. First, we fix the kind of inference
queries we will be using, following rather standard practices. Our
queries will be of the form:
\begin{equation}
\label{eqn:query}
Q\big(\mathcal{B}, on, en_{1}:p_{1}, \ldots, en_{n}\colon p_{1}\big)
\end{equation}
\noindent where:
\begin{itemize}
\item $\mathcal{B}$ is the Bayesian network that the query is applied
  to;

\item $on$ is the `observation node';

\item $en_{1}, \ldots, en_{n}$ are evidence nodes, with predicates
  $p_{1}, \ldots, p_{n}$ on the underlying sets of these nodes.
\end{itemize}
\noindent In the current context we allow the predicates to be fuzzy
(soft); this goes beyond existing approaches.

Our inference algorithm, represented in~\eqref{eqn:query} above as a
query $Q$, performs the following consecutive steps.
\begin{enumerate}
\item Stretch the Bayesian network $\mathcal{B}$ to a linear chain of
  channels, as described in the previous subsection.

\item Locate the observation node and the evidence nodes at the right
  stages in this chain, where these nodes appear for the first time as
  codomain (outcome) of the corresponding channel. The resulting
  situation is sketched in Figure~\ref{fig:transformation}.

\item \label{alg:statetransf} Perform state transformation from the
  initial node to the observation node, while updating the state with
  evidence along the way.

More concretely, take as first state $\omega_{1} = \textsf{initial}$.
Given state $\omega_{i}$ at stage $i$ with subsequent channel $i$,
take:
\[ \begin{array}{rcl}
\omega_{i+1}
& \coloneqq &
\left\{\begin{array}{ll}
c_{i} \gg \big(\omega_{i}|_{p}\big) \quad &
   \mbox{if there is predicate $p$ at stage $i$ in the chain}
\\
c_{i} \gg \omega_{i} & \mbox{otherwise.}
\end{array}\right.
\end{array} \]
\noindent We continue doing this until the `observe' node point is
reached in the chain, from above. The state obtained so far is called
$\omega$.

(There is a subtlety that the predicate $p$ should be suitably
weakenend to be of the right type. The same holds for the predicate
$p$ in the next point; we ignore these matters at this stage; a
precise description is given in Example~\ref{ex:asiachaininfer}
below.)

\item \label{alg:predtransf} Perform predicate transformation from the
  final node to the the observation node, accumulating evidence along
  the way.

More concretely, take as predicate $q_{n} = \one$, where $n$ is the
length of the chain. If there happens to be evidence with predicate
$p$ at the last stage, we take $q_{n} = p$ instead. Assume
predicate $q_{i+1}$ is given at stage $i+1$. A new predicate $q_{i}$ is
formed for the previous stage, via:
\[ \begin{array}{rcl}
q_{i}
& \coloneqq &
\left\{\begin{array}{ll}
\big(c_{i+1} \ll q_{i+1}\big) \andthen p \quad &
   \mbox{if there is predicate $p$ at stage $i$ in the chain}
\\
c_{i+1} \ll q_{i+1} & \mbox{otherwise.}
\end{array}\right.
\end{array} \]
\noindent (We recall that $\andthen$ is used for pointwise
multiplication of fuzzy predicates.) We continue doing this until we
reach the observation node, this time from below. Write $q$ for the
predicate that has been built up at that point.

\item \label{alg:return} Return the updated state $\omega|_{q}$,
  marginalised appropriately to node $on$. This is the output of the
  inference algorithm. Marginalisation is needed because the
  underlying set of the state $\omega$ is typically a product of
  several spaces.
\end{enumerate}
In the code fragment below we illustrate how this looks like for the
query that we considered in the previous section: dyspnea likelihood
given presence in Asia and negative xray. First we load the Asia
Bayesian network from the \texttt{bnlearn} library, in bif format, and
turn it into a Bayesian model in \texttt{pgmpy}.
\begin{python}
>>> reader = BIFReader('asia.bif')
>>> asia_model = reader.get_model()
\end{python}
Our new algorithm's answer to the inference query is produced via the
function \pythoninline{stretch_and_infer}, with appropriate arguments:
\begin{python}
>>> stretch_and_infer(asia_model, 'dysp', 
...   {'asia' : [1,0], 'xray' : [0,1]})
0.3669|dysp_0> + 0.6331|dysp_1>
\end{python}
For comparison, using \texttt{pgmpy}'s variable elimination one writes:
\begin{python}
>>> asia_inference = VariableElimination(asia_model)
>>> asia_inference.query(['dysp'], 
...   evidence={'asia' : 0, 'xray' : 1})['dysp'] )
\end{python}
The latter produces the same distribution, but pretty-printed differently:

\includegraphics[scale=0.4]{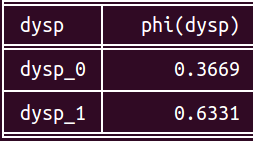}

%% \begin{verbatim}
%% ╒════════╤═════════════╕
%% │ dysp   │   phi(dysp) │
%% ╞════════╪═════════════╡
%% │ dysp_0 │      0.3669 │
%% ├────────┼─────────────┤
%% │ dysp_1 │      0.6331 │
%% ╘════════╧═════════════╛
%% \end{verbatim}

As mentioned, our implementation allows the use of fuzzy, non-sharp
predicates, expressing uncertainly about the evidence at hand, as in:
\begin{python}
>>> stretch_and_infer(asia_model, 'dysp', 
...   {'asia' : [0.9, 0.2], 'xray' : [0.1, 0.75]})
0.3711|dysp_0> + 0.6289|dysp_1>
\end{python}

\begin{exmp}
\label{ex:asiachaininfer}
We also describe more concretely how inference along a chain of
channels works in the new algorithm. We use the same query as above,
applied to the chain of channels described in
Figure~\ref{fig:asiachain}. We use the language EfProb.  First, a
state \pythoninline{forward} is defined via state transformation
$\gg$, as in step~\eqref{alg:statetransf} in the above description of
the algorithm, where the evidence \pythoninline{tt} of having been in
asia is incorporated immediately at the first step. The observation
node is after the \pythoninline{dysp} channel, so that we can follow
the chain in Figure~\ref{fig:asiachain} in a step-by-step manner:
\begin{python}
>>> forward = (dysp @ id) 
...           * (id @ copy) 
...           * (id @ either) 
...           * (id @ id @ tub) 
...           * (id @ lung @ id) 
...           * (bronc @ id @ id) 
...           * (copy @ id) 
...           >> ((smoke @ asia) / (one @ tt))
\end{python}
Notice that the evidence \pythoninline{tt} (of having been in asia) on
the set $2$ is weakened to a predicate \pythoninline{one @ tt} on
$2\times 2$, so that the types match.

Similarly, a predicate \pythoninline{backward} is defined by starting
at the end of the chain, as in step~\eqref{alg:predtransf} of the
algorithm. In this case the negative-xray evidence \pythoninline{ff}
is present at the last point of the chain, and only one predicate
transformation step $\ll$ is needed to reach the observation point
(after \pythoninline{dysp}), see again Figure~\ref{fig:asiachain}.
\begin{python}
>>> backward = (id @ xray) << (one @ ff)
\end{python}
We are now ready to produce the outcome, in step~\eqref{alg:return} of
the algorithm, via state update and marginalisation (using state
transformation with the first projection \pythoninline{pi1}):
\begin{python}
>>> pi1 >> (forward / backward)
0.3669|t> + 0.6331|f>
\end{python}
\end{exmp}

In principle, the stretch step of our inference algorithm can be
executed once, as a precomputation. Subsequent multiple queries can
then be run on this stretched network. However, in the next subsection
we discuss an optimisation step which is query-dependent and make such
pre-computation pointless.

\subsection{Optimisation of the inference algorithm}\label{subsec:optim}

The implementation that we have developed for the above inference
algorithm is a prototype. Squeezing out the last cycle for efficiency
has not been a design goal. Nevertheless, two optimisations have been
included, which improve the algorithm as sketched in the previous two
subsections.

\subsubsection{Doing dry runs first}\label{subsubsec:dryrun}

As we described in Subsection~\ref{subsec:stretch}, stretching of a
Bayesian network is a non-deterministic process. From a mathematical
perspective, it does not matter which order of channels appears in the
resulting chain. However, from a computational perspective it is
important to keep the `width' of the chain as small as possible.  This
width is defined as the maximum size of intermediate product sets in
the chain. For instance, in the stretching in
Figure~\ref{fig:asiachain} the maximal width is 8, given by the
8-element set $2\times 2\times 2$. Different orderings of channels in
a chain can lead to different widths.

%% We illustrate this concretely for the chain: [smoke, asia,
%%   bronc, lung, tub, either, dysp, xray], for the Asia network,
%% mentioned in the beginning of Subsection~\ref{subsec:stretch}.
%% \begin{itemize}
%% \item After adding the initial nodes smoke, asia we have $2\times 2$
%%   as domain.
%% \item Next, for bronc we use the `smoke' part, but we need to keep a
%%   copy of smoke for later use; hence the domain is now $2\times
%%   2\times 2$.
%% \item The lung channel consumes the second `smoke' part, but
%%   introduces a new part; so we remain at $2\times 2\times 2$.
%% \item Similarly, after tub we are still at $2\times 2\times 2$.
%% \item The either node has two parents, and thus consumes two $2$',
%%   yielding output $2\times 2$.
%% \item When we apply dysp consume two $2$'s, but we need to keep a copy
%%   of the either output for the xray channel. Hence we remain at
%%   $2\times 2$.
%% \item Finally, the xray channel consumes one $2$, but produces a new
%%   one, so the final domain is $2\times 2$.
%% \end{itemize}
%% %
%% We see that the width of this channel is $\#2\cdot\#2\cdot\#2 = 8$.
%% This is not very much, but for bigger networks, with multiple (more
%% than 2) items in domains, this width grows quickly. Hence it is
%% important to select the most economical channel.

What our implementation does is perform a number of random `dry'
stretch runs, say 1000, to find out in a brute force manner which of
these chains has the least width. This is the one that we proceed
with, in order to fill in the precise `burocratic' details about
coping and swapping in order to line up inputs appropriately for each
channel. As an illustration, doing 100 dry runs on the Child network
(in the middle of Figure~\ref{fig:childstretch}) yields the following
28 different possible widths (duplicate occurrences have been removed).
\[ {\renewcommand{\arraystretch}{1.0}\begin{array}{lclclcl}
19440 
& \qquad &
9720
& \qquad &
38880
& \qquad &
4860
\\
7776
& &
3888
& &
15552
& &
12960
\\
6480
& &
3240
& &
58320
& &
8640
\\
23328
& &
46656
& &
25920
& &
5184
\\
11664
& &
5832
& &
3456
& &
77760
\\
2592
& &
51840
& &
4320
& &
116640
\\
93312
& &
3456
& &
11520
& &
17280
\end{array}} \]

\noindent The smallest width --- 2592 in this case --- is selected for
the remainder of the algorithm. Doing such dry runs is computationally
cheap, and well worth spending a little bit of time on, since the
variation in widths is substantial.

Instead of these brute force `dry' runs one could use some more clever
graph analysis techniques. Such improvements may be added at a later
stage; they are besides the main focus of this paper, namely the
application of state/predicate transformer semantics in inference.

\subsubsection{Pruning the Bayesian network first}\label{subsubsec:prune}

An inference query as in~\eqref{eqn:query} contains several nodes,
namely for observation ($on$) and for evidence ($en_i$). These nodes
occur at specific parts in the Bayesian network $\mathcal{B}$ at
hand. This means that there are parts of the network which are
irrelevant for the inference query. Specifically, if there is a node
$n$ in the network and neither the observation node $on$ nor any of
the evidence nodes $en_i$ occurs in the subnetwork of children of $n$
(including $n$ itself), then $n$ may as well be removed. This is what
we apply in our inference algorithm, to the \texttt{pgmpy} model,
before it is stretched\footnote{Via the method
  \pythoninline{remove_node} of the \pythoninline{BayesianModel}
  class.}.

% http://pgmpy.org/models.html#module-pgmpy.models.BayesianModel

This `pruning' of the Bayesian network, by removing irrelevant nodes,
greatly improves the efficiency. But it makes the whole inference
algorithm query dependent. Hence the idea that the chain of channels
can be pre-computed does not work anymore with this optimisation step.

\subsection{Comparison with variable elimination in 
   \texttt{pgmpy}}\label{subsec:comparison}

The aim of the prototype implementation of our channel-based inference
algorithm is mainly to see if the idea of reasoning up and down a
chain of channels works. We can surely say that it does, from a
functional perspective. We have compared it to the standard variable
elimination algorithm for inference, as implemented in
\texttt{pgmpy}. In all our tests, the outcomes have been the same.

We have also done some performance comparisons. The results described
below do give some indication, but not much more than that. For
instance, we have only compared to variable elimination in
\texttt{pgmpy}, and not to other implementations. Also, the comparison
is complicated by the fact that our inference algorithm is
non-deterministic --- and it seems, variable elimination in
\texttt{pgmpy} too. Hence the only thing we can do is compare many
runs, on different queries. Here are some findings.
\begin{itemize}
\item On small examples, like Asia from Section~\ref{sec:baynet},
  there is no noticeable timing difference. This Asia model involves 8
  nodes and 18 parameters\footnote{We follow the counting from
    \url{bnlearn.com/bnrepository}.}.

\item On examples like the Child network from
  Figure~\ref{fig:childstretch}, with 20 nodes and 230 parameters,
  there are clear differences. We have run many randomly generated
  queries, with between 1 and 5 evidence nodes. On average, on this
  example, our algorithm is in the order of 10 times faster, on an
  ordinary laptop. Even with larger numbers of runs, substantial
  variation in execution times remain\footnote{The script we use will
    be made available on \url{efprob.cs.ru.nl} so that people can
    check and try for themselves.}. This variation occurs in
  particular for variable elimination, not for the new algorithm.

%% Hoi Bart,
%% Goede omschrijving van de server is:
%% "Dell R920 (quad processor E7-4870v2 at 2.3 Ghz, with 60/120 cores/threads 
%% total, with 3 TiB RAM at 1600 Mhz), running Ubuntu 16.04 LTS".
%% Groetjes,  Bernard

\item For larger Bayesian networks from \texttt{bnlearn}, like
  Insurance with 27 nodes and 984 parameters, our inference algorithm
  often takes about a second to terminate, whereas variable
  elimination in \texttt{pgmpy} is typically hundreds or even
  thousands of times slower. We have done these experiments both on a
  laptop and on a quad processor machine with 3 TiB RAM.

\item For even larger networks, like Hailfinder with 2656 parameters,
  our algorithm fails, because of lack of memory, on the big 3 TiB RAM
  machine.
\end{itemize}

%% Child experiments:

%% Total inference time is:  1049.0210511226323  for  1000  runs
%% of which for variable elimination:  1013.7814821136999
%% and for transformations:  35.23956900893245
%% How much faster is transformations inference: 28.768271310489887

%% Total inference time is:  736.9502224680618  for  1000  runs
%% of which for variable elimination:  701.5042858982342
%% and for transformations:  35.44593656982761
%% How much faster is transformations inference: 19.790823823099956

%% Total inference time is:  95.76268152793637  for  1000  runs
%% of which for variable elimination:  52.47815000626724
%% and for transformations:  43.28453152166912
%% How much faster is transformations inference: 1.212399630107943

%% Total inference time is:  269.6872955505969  for  1000  runs
%% of which for variable elimination:  233.85500987572595
%% and for transformations:  35.832285674870946
%% How much faster is transformations inference: 6.526377133673268

%% Total inference time is:  381.11831351555884  for  1000  runs
%% of which for variable elimination:  343.2624232598464
%% and for transformations:  37.85589025571244
%% How much faster is transformations inference: 9.067609318949994

\section{Irrelevance of channel ordering}\label{sec:correct}

In subsection~\ref{subsec:stretch} we have made a casual remark about
stretching a Bayesian network, namely that from a mathematical
perspective, the order of channels in a chain does not matter.
Actually, there is something to prove here, when we do conditioning.
Consider the situation sketched in Figure~\ref{fig:transformation}.
It may happen that in one ordering of channels a particular piece of
evidence (predicate) is \emph{above} (before) the observation point,
whereas in another ordering it is \emph{below} (after). According to
steps~\ref{alg:statetransf} and~\ref{alg:predtransf} in
Subsection~\ref{subsec:transform} this predicate will then be treated
differently: if it occurs \emph{above}, the predicate will be
incorporated in the final outcome via \emph{state} transformation; if
it occurs \emph{below}, the predicate will be handled via
\emph{predicate} transformation. We need to show that this yields the
same outcomes.

Emperically, we know from running the algorithm multiple times that
the outcome is independent of such orderings. But we better prove this
mathematically. The theorem below abstracts the situation to its
essential form.

\begin{thm}
\label{thm:correct}
Let $c\colon \kto{A}{X}$ and $d\colon \kto{B}{Y}$ be channels,
together with a joint state $\omega\in\Dst(A\times X)$ and a predicate
$q\colon Y\rightarrow [0,1]$ on $Y$. Then the following distributions
are the same.
\begin{equation}
\label{eqn:correct}
\begin{array}{rcl}
\lefteqn{\pi_{1} \gg \Big(\big((c\otimes\idmap) \gg \omega\big)
   \big|_{(\idmap\otimes d) \ll (\one\otimes q)}\Big)}
\\
& = &
\pi_{1} \gg \Big((c\otimes\idmap) \gg 
    \big(((\idmap\otimes d) \gg \omega)\big|_{\one\otimes q}\big)\Big).
\end{array}
\end{equation}
\end{thm}

This equation deserves some explanation. The outer operation $\pi_{1}
\gg (-)$ performs marginalisation, for observing the outcome in $X$.
In the upper expression the channel $d$ is used for predicate
transformation \emph{after} using $c$ for state transformation.  In
the lower expression $d$ is used for state transformation,
\emph{before} applying $c$, also for state transformation, to an
updated state. In the style of Figure~\ref{fig:transformation}, the
upper expression in~\eqref{eqn:correct} captures the situation on the
left below, whereas the lower expression in~\eqref{eqn:correct} is
about the picture on the right:
\[ \xymatrix@C-1pc{
& A\times B\ar|{\circ}[d]_{c\otimes\idmap}
& &
& A\times B\ar|{\circ}[d]_{\idmap\otimes d}
\\
\mbox{\textcolor{blue}{observe $X$}}\ar@{=>}[r] & 
   X\times B\ar|{\circ}[d]_{\idmap\otimes d} 
& &
& A\times Y\ar|{\circ}[d]_{c\otimes\idmap} & 
   \mbox{\textcolor{red}{evidence on $Y$}}\ar@{=>}[l]
\\
& X\times Y & \mbox{\textcolor{red}{evidence on $Y$}}\ar@{=>}[l]
&
\mbox{\textcolor{blue}{observe $X$}}\ar@{=>}[r] & X\times Y
} \]
This shows that channels $c$ and $d$ can be shifted
along each other in a non-interacting manner, in the presence
of conditioning with predicate $q$

\begin{proof}
For convenience we shall identify a distribution $\rho\in \Dst(Z)$
with a function $\rho\colon Z \rightarrow [0,1]$, in a standard
manner: if $Z = \{z_{1}, \ldots, z_{n}\}$ and $\rho$ is
$\sum_{i}r_{i}\ket{z_i}$, then $\rho\colon Z \rightarrow[0,1]$ is
given by $\rho(z_{i}) = r_{i}$. Thus, for $x\in X$,
\[ \begin{array}[b]{rcl}
\lefteqn{\pi_{1} \gg \big(\big((c\otimes\idmap) \gg \omega\big)
   \big|_{(\idmap\otimes d) \ll (\one\otimes q)}\big)(x)}
\\
& = &
\displaystyle\sum_{b\in B} \big(\big((c\otimes\idmap) \gg \omega\big)
   \big|_{(\idmap\otimes d) \ll (\one\otimes q)}\big)(x,b)
\\
& = &
\displaystyle\sum_{b\in B} \frac{((c\otimes\idmap) \gg \omega)(x,b)\cdot
   ((\idmap\otimes d) \ll (\one\otimes q))(x,b)}
   {(c\otimes\idmap) \gg \omega \models 
    (\idmap\otimes d) \ll (\one\otimes q)}
\\
& = &
\displaystyle\sum_{b\in B} \frac{\big(\sum_{a\in A}c(a)(x)\cdot\omega(a,b)\big)
   \cdot\big(\sum_{y\in Y}d(b)(y) \cdot q(y)\big)}
   {(\idmap\otimes d) \gg \omega \models 
     (c\otimes\idmap) \ll (\one\otimes q)}
\\
& = &
\displaystyle\sum_{y\in Y}\sum_{a\in A}\sum_{b\in B} 
   \frac{c(a)(x)\cdot\omega(a,b)\cdot d(b)(y) \cdot q(y)}
   {(\idmap\otimes d) \gg \omega \models 
     (c \ll \one)\otimes (\idmap \ll q)}
\\
& = &
\displaystyle\sum_{y\in Y} \sum_{a\in A} 
    \frac{c(a)(x) \cdot\big(\sum_{b\in B} d(b)(y)\cdot\omega(a,b)\big) \cdot q(y)}
         {(\idmap\otimes d) \gg \omega\models \one\otimes q}
\\
& = &
\displaystyle\sum_{y\in Y} \sum_{a\in A} c(a)(x) \cdot
    \frac{((\idmap\otimes d) \gg \omega)(a,y) \cdot (\one\otimes q)(a,y)}
         {(\idmap\otimes d) \gg \omega\models \one\otimes q}
\\
& = &
\displaystyle\sum_{y\in Y} \sum_{a\in A} c(a)(x) \cdot
    \big(((\idmap\otimes d) \gg \omega)\big|_{\one\otimes q}\big)(a,y)
\\
& = &
\displaystyle\sum_{y\in Y} \big((c\otimes\idmap) \gg 
    \big(((\idmap\otimes d) \gg \omega)\big|_{\one\otimes q}\big)\big)(x,y)
\\
& = &
\pi_{1} \gg \big((c\otimes\idmap) \gg 
    \big(((\idmap\otimes d) \gg \omega)\big|_{\one\otimes q}\big)\big)(x).
\end{array}\eqno{\QEDbox} \]
\end{proof}

\section{Concluding remarks}\label{sec:conclusions}

This paper concentrates on the underlying semantical ideas for a new
algorithm for exact Bayesian inference. It builds on ideas from
(probabilistic) programming semantics, using the notion of channels as
primitive, both for state transformation and predicate
transformation. The comparison of the prototype implementation of the
algorithm in Python to a standard implementation (from \texttt{pgmpy})
gives a first indication, but clearly requires a more systematic
analysis, involving different inference algorithms and different
implementations. This is beyond the focus of the current paper.

Since the approach is based on a very general underlying semantics, it
can be extended in principle also to continuous probability theory.
Categorically, this amounts to using the Giry monad $\Giry$ instead of
the distribution monad $\Dst$, see~\cite{Jacobs17a} for details. Even
more, it could apply to hybrid networks, combinining both discrete and
continuous probability (see~\cite{ChoJ17b} for an example). Even more,
the approach could be extended to quantum Bayesian probability, since
channels (also called superoperators, see~\cite{NielsenC00}
or~\cite{Jacobs15d}) are a primitive notion in quantum computing too.

Another avenue for further work is to apply the stretching techniques
described here also to MAP-inference. This is ongoing work.

\subsection*{Acknoledgements} Thanks are due to Kenta Cho, Fabio Zanasi
and Marco Gaboardi for discussions and feedback.

%\bibliography{/home/bart/svn/bart/Tex/bib}

\end{document}